# Understanding Natural Language Understanding Systems.
# A Critical Analysis

Alessandro Lenci

> "It was religious yearning granted hope, it was the holy grail of science."
> McEwan, *Machines like me*

1. THE DAWNING OF A NEW ERA FOR ARTIFICIAL INTELLIGENCE?

The development of machines that «talk like us», also known as Natural Language Understanding (NLU) systems, is the Holy Grail of Artificial Intelligence (AI), since language is the quintessence of human intelligence. The brief but intense life of NLU research in AI and Natural Language Processing (NLP) is full of ups and downs, with periods of high hopes that the Grail is finally within reach, typically followed by phases of equally deep despair and disillusion (Bowman 2022). But never has the trust that we can build «talking machines» been stronger than the one engendered by the last generation of NLU systems. Indeed, the mastery of *chatGPT*, a conversational agent that represents the most advanced product (until now) of such generation, is nothing but impressive, as these examples below show:[1]

(1) Human question: *John is richer than Mary. Mark has more money than Mary, but less than John. Is John richer than Mark?*
chatGPT: *Yes, John is richer than Mark.*

(2) Human question: *Yesterday evening, John asked Mary to go to the cinema with him. Mary replied: "Tomorrow, you have an exam". What did Mary intend to say?*
chatGPT: *Mary intended to remind John that he has an exam scheduled for the next day, and it may be best for him to focus on studying instead of going to the cinema. She is suggesting that his priority should be his responsibilities and studying for the exam before going out and having fun.*

(3) Human question: *John has 10300 euros, but he owes 4725 to Mary. John wants to change his car and yesterday he saw one that costs 7300 euros. Can John afford to buy that car?*
chatGPT: *John has 10300 euros and owes 4725 to Mary, so he has 10300 - 4725 = 5575 euros available to spend. The car he wants to buy costs 7300 euros, so he cannot afford it with the money he has available. 5575 is less than 7300.*

In answering the human questions, chatGPT carries out a transitive inference in (1), identifies that Mary performed an indirect speech act in (2) and solves a nontrivial mathematical problem in (3), even providing an explanation for it. Moreover, it produces grammatically sound, semantically coherent, and stylistically fluid texts. The performance of NLU systems like chatGPT is indeed unprecedented, like the fact that for the first time they are accessible to the general public, who can freely interact with them in several languages.[2] Therefore, it is quite unsurprising that the abilities of such systems has engendered the impression that they indeed «understand» natural language and manifest human-like intelligent behaviors.

---

[1] chatGPT interrogated on 15 January 2023 at https://chat.openai.com/chat.
[2] This is part of an astute business strategy of Open AI, the developer of chatGPT and other similar NLU systems, to raise media attention for their models, and therefore foster their following use in commercial applications.

The media attention to current NLUs systems is huge and again unprecedented (Dale 2021). In July of 2022, press reported that Google fired one of their engineers for having claimed in an interview with the Washington Post that the LaMDA system (Thoppilan *et al.* 2022) had become sentient.[3] The engineer had conducted several conversations with the system during which he increasingly felt that the model's replies indicated sentience. This is an extreme case, but media worldwide are full of articles that hail these systems and describe their «extraordinary» abilities in strongly humanized terms (Shanahan 2022).[4] Besides the media rhetoric, this tendency is understandable, since it is natural to explain sophisticated linguistic behaviors by ascribing epistemic and intentional states to the entity that produce them.[5] If a human speaker provides the answers in (1)-(3), we would not hesitate to say that she has a very good mastery of English that allows her to understand the questions, that she has semantic (e.g., she knows that being richer than someone means having more money than her) and pragmatic skills (e.g., she understands that in saying *Tomorrow, you have an exam* Mary refused John's invitation), and that she has remarkable mathematical and reasoning abilities. Therefore, if a machine generates the same answers, it seems straightforward to conclude that this is because it has similar knowledge and abilities.

Another striking fact of nowadays NLU systems is their *prima facie* creativity and generality. Previous models (rule-based ones, but even those based on machine learning algorithms) were still very limited in the type of lexical and grammatical structures they could master and in the intelligent abilities they showed, which were essentially restricted to the specific tasks they were trained for. The capabilities of state-of-the-art systems are instead apparently open-ended: chatGPT answers questions virtually on any topic, translates, generates scientific texts as well as informal ones, produces and understand jokes, poems, and metaphors, debates on abortion or racism, and so on. It is this versatility that contributes to the impression that such systems are approximating the hallmark of human language understanding: *A general ability that allows us to understand and produce a potentially unlimited number of utterances and master any possible communicative situation*.

But is it gold all that glitters in AI? do systems like chatGPT possess something comparable to the human knowledge of language? Are we at the dawn of a new era, in which the Grail is finally closer to us?[6] In fact, the latest achievements of AI systems have sparkled, or better renewed, an intense scientific debate on their true language understanding capabilities. Some defend the idea that, yes, we are on the right track, despite the limits that computational models still show. Others are instead radically skeptic and even dismissal: The present limits are not just contingent and temporary problems of NLU systems, but the sign of the intrinsic inadequacy of the epistemological and technological paradigm grounding them. This paper aims at contributing to such debate by carrying out a critical analysis of the linguistic abilities of the most recent NLU systems. I contend that they incorporate important aspects of the way language is learnt and processed by humans, but at the same time they lack key interpretive and inferential skills that it is unlikely they can attain unless they are integrated with structured knowledge and the ability to exploit it for language use.

The paper is organized as follows. Section 2 reviews the main features of current NLUs models, how they learn and represent their «knowledge» (I put epistemic and intentional terms between quotes when used to refer to their states, i.e., human knowledge of language vs. chatGPT's «knowledge» of language). Section 3 recapitulates the most important positions and arguments in the debate about NLUs, and focuses on the key question of the paper: *To what extent the way*

---

[3] N. Tiku, «The Google engineer who thinks the company's AI has come to life». *The Washington Post*, June 21, 2022.
[4] For instance, in the Italian newspaper *La Repubblica* (December 14, 2022) Massimo Rocca, after playing with chatGPT, says that «it is a compassionate and progressist friend, who loves to improvise rap verses». Of course, Massimo Rocca does not believe that the system is sentient, but anyway he talks about chatGPT as it were so.
[5] I also have found extremely hard to refrain myself from using words like believe or intend to describe chatGPT's answers in (1)-(3).
[6] For instance, the *Bloomberg* editorial on 9 December 2022 does not hesitate to define chatGPT as the «start of the AI revolution», which is expected to have deep social, economic, and even ethical impacts.

*machines «understand» and «use» natural language approximates the way humans understand and use it to communicate among them, and what is still missing to fill the gap.*

2.  HOW DO STATE-OF-THE ART NLU SYSTEMS «KNOW» WHAT THEY «KNOW»?

Last generation NLU systems all share the same architecture: Deep artificial neural networks *pretrained* on huge amounts of unlabeled textual data, that are then adapted to a wide range of downstream tasks. These *foundation models*, as they are referred to by Bommasani *et al.* (2021), represent a genuine breakthrough with respect to the previous paradigm in AI and NLP, which was based on a «one task, one model» approach, consisting in training a distinct model for each specific application (e.g., syntactic parsing, question answering, machine translation, etc.), typically using a supervised machine learning algorithm. The bottleneck of such method was represented by the need of large quantities of labelled data (e.g., sentences annotated with their syntactic structure), which are notoriously long and expensive to produce, and its lack of generality and limited adaptability to new tasks and domains (Manning 2022). Foundation models are instead closer to the view that human cognition is endowed with general-purpose knowledge of language and of the world, which is used to perform various communicative tasks, from interpreting a joke to answering a question, and can be adapted and extended to tackle new domains or situations. For instance, somebody who knows Italian and starts reading legal texts, does not need to learn Italian from scratch, but can exploit her knowledge of this language to adapt it to the peculiarities of the legal domain (e.g., by acquiring new words and grammatical constructions).

The power of foundation models mainly resides in the pretraining phase, which allows the network to acquire a large amount of «core knowledge» by being trained as a *language model*. In fact, foundation models are also commonly called *pretrained neural language models* (Han *et al.* 2021). To know what a language model is, let's ask to a language model itself:

> *A language model is a type of machine learning model that is trained to generate text or predict the next word in a sequence of words. It is used in various natural language processing tasks such as language translation, text summarization, and speech recognition. Language models are trained on large amounts of text data and use statistical techniques to learn the patterns and structure of the language. They can be used to generate new text or to make predictions about the likelihood of different words or phrases appearing in a given context.*[7]

Thus, a neural language model is an artificial neural network trained on a task of string prediction (Goldberg 2017). State-of-the-art foundation models are based on Transformers (Vaswani et al. 2017), multi-layered networks that have *de facto* replaced previous architectures, like recurrent ones. The key aspect of language modeling is that it does not require annotated data, because the learning objective of the network is to reproduce the input word sequence itself, hence the name of *self-supervised leaning* (Manning, Clark, Hewitt, Khandelwal and Levy 2020). There are two major types of language models, depending on the kind of prediction task:
i.) *autoregressive (or causal) language models*, like the members of the GPT family (Generative Pre-trained Transformer; Radford, Narasimhan, Salimans and Sutskever 2018, Radford *et al.* 2019, Brown *et al.* 2020), are unidirectional and trained to predict the conditional probability of a target word $w_n$ (e.g., *ball*) given the preceding context $w_1, ..., w_{n-1}$ (e.g., *The dog is chasing a red ...*). These language models are also called *generative*, because they are optimized to generate the most likely text sequences following a ceratin context;

---

[7] This is the full answer provided by chatGPT to the question *What is a language model?*

ii.) *denoising autoencoding models*, like BERT (Bidirectional Encoder Representations from Transformers; Devlin, Chang, Lee and Toutanova 2019), are trained on a *masked language modeling* task inspired to the Cloze test: Some of the tokens from the input are randomly masked (e.g., *The dog is [MASK] a red ball*), and the network training objective is to predict the original masked words based on their context. These language models are bidirectional, because a word is predicted by considering both the preceding and the following items.

Foundation models acquire their «core knowledge» as a side effect of learning to predict lexical items given a certain context. Therefore, this «knowledge», encoded in the vectors corresponding to the internal layers of the model, contains *only* the information that can be recovered from the co-occurrence statistics in the input texts (e.g., that *ball* tends to co-occur with *kick* or *chase*), implicitly recorded though the prediction training task. The relationship between meaning and linguistic co-occurrences is the focus of *distributional semantics* (Lenci 2008, 2018; Lenci and Sahlgren 2023), an approach to meaning that develops theories and methods for representing and acquiring semantic properties of linguistic items from their distributional properties in text corpora. Grounded in the so-called *Distributional Hypothesis*, according to which words with similar linguistic contexts tend to have similar meanings, distributional semantics represents the meaning of a linguistic expression with a real-valued vector (nowadays commonly called *embedding*) that encodes its statistical distribution in contexts. The continuous nature of its representations distinguishes distributional semantics from other theoretical frameworks that instead represent meaning with symbolic structures. *Distributional semantic models* (DSMs) are computational methods to learn semantic representations from corpus data, which are nowadays widely used in NLP and AI applications, as well as for cognitive and linguistic analyses. Neural language models have become the most popular method to learn word embeddings at least since the advent of *wod2vec* (Mikolov, Chen, Corrado and Dean 2013), which has started the generation of so-called *predict DSMs* that have superseded previous models based on the explicit counting of distributional data in corpora and their representation with co-occurrence matrices.

Foundation models have evolved from predict DSMs into a radically new generation of methods to acquire knowledge from texts, with the following main elements of novelty:

i.) *the type of representations* they learn. Traditional DSMs are essentially models of the lexicon, intended as a repository of out-of-context lexical items. Each word type in the model vocabulary is represented with a unique, context-independent, *static embedding*, which thus conflates different word senses in the same vector. Foundation models instead take in input a whole sentence or discourse and generate a *contextual embedding* for each of its word tokens (Lenci and Sahlgren 2023). This means that the same lexical item in different contexts (e.g., *bat* in *A bat is flying in my room* and in *I hit the ball with a bat*) will be represented with distinct embeddings. The context-sensitive nature of their representations is one of the reasons of the much-improved performance of foundation models, as they are able to capture fine-grained aspects of word senses;

ii.) *the model size*, which has been characterized by an exponential growth both in their architecture and in the amount of training text. This justifies the name of *large language models*. The large version of BERT contains 24 layers and 340 million parameters (i.e., the weights that are set during training and determine the network behavior). GPT-3, which is the foundation model used to develop chatGPT, has 96 layers and 175 billion parameters, an increase of several order of magnitudes in just two years. The size of training data for GPT-3 is equally huge, about 499 billion tokens. To have an idea of what this means, consider that the whole Wikipedia is just the 3% of the training texts and that the average data size of the previous generation of predict DSMs was around three billion tokens. Obviously, this also means an enormous increase in the computational resources necessary to train foundation models, which has in fact raised the issue of their sustainability (Strubell, Ganesh and McCallum, 2019);

iii.) *the amount of information* they encode. Thanks to the context-sensitive nature of their representations and the size of their networks and training data, foundation models learn from texts a far greater amount of information than any former DSM. Contextual embeddings capture aspects of syntactic structure, several dimensions of lexical and sentence meaning, and so on (Tenney *et al.* 2019; Manning, Clark, Hewitt, Khandelwal and Levy 2020; Rogers, Kovaleva and Rumshisky 2020; among several others). Perhaps even more strikingly, the largest models like GPT-3 reveal «emergent abilities» to carry out linguistic tasks (e.g., translating, question-answering, etc.) without any task-specific training (Brown *et al.* 2020, Wei *et al.* 2022). Therefore, the «core knowledge» of foundations models is not only a «knowing that» of structures of languages and facts of the world, but also a «knowing how» to use language itself.

A further key innovation of foundation models is the way their «core knowledge» is used to develop downstream NLU applications. The common approach consists in adapting the foundation model to different tasks via *fine-tuning*: The pretrained model is re-trained on a specific supervised task (e.g., question-answering) to adapt its weights to perform the new task. Since the model can leverage the «core knowledge» encoded during pretraining, excellent performances can be achieved with smaller amount of labeled data than the ones required by traditional machine learning methods. The «magic» of chatGPT is exactly the product of such a paradigm. This model is a further refinement of instructGPT (Ouyang *et al.* 2022), which is GPT-3 fined-tuned via reinforcement learning using human feedback on the texts it generates, with the purpose of reducing some of its major limitations, such as the generation of inappropriate texts, loss of discourse coherence over long passages, contradictory answers, *non sequitur* sentences, and so on. Therefore, it is essential to keep in mind that behind the impressive performances of chatGPT there is actually a large human effort to manually label and correct the texts created by the foundation model. Henceforth, I will focus only the «core knowledge» acquired by foundation models during pretraing, leaving aside the possible further improvements that can be obtained via fine-tuning.

Very large models like GPT-3 have also opened the way to a new unsupervised method of task adaptation called *in-context* or *zero-shot learning*. This consists in being able to solve tasks that the model has not been trained for, by taking advantage of the «emergent abilities» of the foundation model itself (see above), and simply giving it a carefully constructed instruction as input sequence, called *prompt*, that contains a natural language description of the task. For instance, if we want to do machine translation from Italian to French, we will simply give to GPT-3 the prompt *translate from Italian to French:* and the input the text we want the model to translate.

In summary, how do state-of-the art NLU systems «know» what they «know»? They «know» it through the distributional analysis of the texts they are pretrained on (see below for foundation models that use non-linguistic data too). How do they represent this «knowledge»? It is encoded in continuous representations corresponding to the embeddings of the network internal states.

3.  LANGUAGE UNDERSTANDING IN MACHINES AND HUMANS

The ability of NLU systems based on foundation models to generate *human-like language* is surely astounding. Therefore, the celebration they are receiving in the media is unsurprising. One recent article reported an experiment carried out by a group of students at an Italian university who created a whole issue of their magazine with chatGPT: despite some few exceptions, the students' and journalists' comments were all enthusiastic.[8] There is in fact a widespread sensation that we are witnessing something radically new, which has determined the resurgence of questions that have dominated the collective imaginary of AI, such as: *Are NLU systems now able to «understand» language? Will they develop a «conscience»? Do they «mean» what they say? Are they «intelligent» in the same sense of the term we apply to humans?*[9] It is curious that questions like these have never been raised about last-generation machine translation systems like Google

---

[8] *La Repubblica*, 29 January 2023.
[9] Attention to the ethical and social issues raised by these NLU systems is equally strong, for instance for their forecast impact on the educational system, misinformation, document forgery, and so on.

Translate, which have also reached a professional quality in translating many types of texts. On the one hand, for humans translating entails understanding the source text they translate (I cannot translate from Chinese, unless I know Chinese). On the other hand, we also know that machine translation systems learn to translate by simply being exposed to large amounts of translation examples derived from parallel corpora. In fact, neural machine translation systems are just an instance of the same technological paradigm described in Section 2 that grounds chatGP, but nobody really thinks they have any real human-analogous understanding of the content of the texts they translate. They are only very good at exploiting and generalizing the cross-lingual distributional correlations they find in the training corpora. NLU systems like chatGPT do not only translate, but also interact with us in a natural way and on a multitude of different tasks and topics. This gives us the impression that there is «something more» in them, but there is not.

The AI and NLP research communities have been equally impressed and stormed by foundation models. Instead of the «metaphysical» questions about conscience and mind that are nowadays common in the media, the scientific debate concerns whether the new paradigm grounding the latest achievements of NLU systems will really lead us closer to develop machines endowed with genuine, human-analogous language understanding. Bender and Koller (2020) aptly formulate this doubt in the question: *are we climbing the right hill?* If the improvements made possible by foundations models are indeed undeniable, equally certain is the fact that they perform several mistakes in terms of both language processing and reasoning («chatGPT error hunting» may indeed soon become a popular party game). Closer analyses reveal that their linguistic behavior is far and, under several respects, extremely different from human one. Thus, the general impression is that we are not quite there yet. There is still something missing, and the key question is understanding what prevents machines from really «talking like us».

3.1. INTELLIGENT MACHINES OR STOCHASTIC PARROTS? THE CURRENT DEBATE ON NLU SYSTEMS

While nobody denies the limits of foundation models, it is often emphasized that they are still in their infancy and much needs to be understood about what they learn and how they internally represent and use what they acquire from texts (actually, this goal is made extremely hard to achieve by the huge complexity and size of the models themselves). Since increasing the size of the models have generated «emergent abilities» to carry out linguistic tasks, this prompts the idea that «additional scaling could further expand the range of capabilities of language models» (Wey *et al.* 2022). According to this view, the existing gap between NLU systems and human language understanding might be mostly a matter of *scale*. Improvements are also to be expected by refining the pretraining methods, as well as their adaptability to new tasks (Yogatama *et al.* 2019, Bommasani *et al.* 2021, Manning 2022). Therefore, the core paradigm of foundational models, based on distributional learning and on non-symbolic representations, would be however substantially correct, and we are «climbing the right hill». On the other hand, skeptical positions have also emerged, which instead target essential aspects of foundation models and the new mainstream approach to NLU.

A first critique concerns the fact that **(1)** foundational models lack access to the extralinguistic world. I will call this the *grounding argument* because it is closely related to the *symbol grounding problem* by Harnad (1990). Bender and Koller (2020) use a variant of the *Chinese Room Argument* by Searle (1980)[10] to claim that models that only observe the co-

---

[10] Bender and Koller (2020) call their argument the *Octopus Test*. A hyper-intelligent octopus discovers a way to listen to the conversations between two individuals and learns English only through the statistical analyses of the signals it detects, like a language model. The octopus becomes so fluent that it begins to intervene in the conversations between the two humans and to generate sentences that «look» meaningful (e.g., because they are perfectly coherent answers to questions). Still, no matter how good its interactions with the human speakers are, the octopus would have no idea of

occurrences of linguistic expressions will never be able to learn meaning, since meaning is inherently a relation between a linguistic form and a communicative intent about *something (either concrete or abstract) external to language*. Therefore, a foundation model like GPT-3 is just «a system for haphazardly stitching together sequences of linguistic forms it has observed in its vast training data, according to probabilistic information about how they combine, but without any reference to meaning: a stochastic parrot» (Bender, Gebru, McMillan-Major and Shmitchell 2021: 617). Bisk *et al.* (2020) also argue that the greatest limit of foundation models resides in their learning only from linguistic data, while humans acquire language though their experiences in the world, via grounding, embodiment, and social interaction. A similar view is expressed by Lake and Murphy (2021), who regard even the latest and most sophisticated DSMs as quite far from being psychologically plausible models of meaning, because they are not connected with perception, action, and reasoning.

As pointed out in Section 2, foundations models can be regarded as the last generation of DSMs and the grounding argument is not new in the history of distributional semantics (Lenci and Sahlgren 2023). Since the appearance of the first generation of DSMs, like Latent Semantic Analysis (Landauer and Dumais 1997), several proponents of the embodied cognition paradigm have expressed deep skepticism about the cognitive plausibility of distributional representations, exactly because of their lack of extralinguistic grounding (Glenberg and Robertson 2000). However, although it is true that the word embeddings produced by DSMs are ungrounded, *this does not entail that they cannot be grounded*. In fact, distributional approaches are not necessarily constrained to textual input. Research in distributional semantics has led to the creation of *multimodal DSMs* that acquire knowledge by combining information from different types of signals, like texts and images. Such combinations have also been the locus of recent developments in multimodal foundations models that can produce text from images, such as CLIP (Radford *et al.* 2021), and models that can produce images from text, like DALL·E 2 (Ramesh, Dhariwal, Nichol, Chu and Chen 2022). These models eschew the grounding problem at least in the sense of combining different modalities, and as such they do reach the «world outside of language». This represents a key difference with the grounding problem that affected classical symbolic AI. In that case, the conundrum was how to establish a mapping between two completely different representation formats: categorical amodal symbols representing concepts (e.g., DOG as the meaning of *dog*), and continuous modal information about the entities these symbols refer to (e.g., images of dogs we have experienced). On the other hand, DSMs and foundation models can be grounded exactly because they represent all kinds of information with *continuous vectors*, which can thus integrate signals coming from textual co-occurrences and from extralinguistic data like images, videos, and so on. Therefore, the grounding problem advanced by Bender and Koller (2020) is not a truly «killing argument» against foundation models, but at most an important encouragement to focus research on their integration with multimodal data, which has already made important progress, as DALL·E 2 shows.

Another limit of the grounding argument is the fact that it relies on a quite narrow view of meaning, inherently defined as a relation between a linguistic form and the world. Although referential aspects are undoubtedly important, still they are not exhaustive of what meaning is. Marconi (1997) introduces a distinction between *referential* and *inferential competence* of word meaning. The former is the ability to relate words to the world, while the latter is knowledge about how words relate to other words (e.g., knowing that *dog* is semantically similar to *puppy*, is a hyponym of *mammal*, etc.). These two kinds of competences are interrelated, but also distinct, even at the neurocognitive level (Calzavarini 2017). For example, we may know several things about the meaning of *aardvark* (e.g., that it is a nocturnal mammal native to Africa), without being able to refer to one. Conversely, someone could name by *kangaroo* the right kind of animal, without knowing that it is related to Australia or is a kind of marsupial. Piantadosi and Hill (2022) defend a

---

what the sentences it receives or produces mean, because it cannot link those sentences to entities and actions in the world.

similar position by adopting a *conceptual role* view of meaning, according to which the meaning of a lexical item comes from its relations with other elements in some wider «theory» (e.g., the meaning of the term *entanglement* derives from the «role» it has within the system of concepts that constitutes the theory of quantum physics). It is also worth remarking that the most recent developments in cognitive science are revisiting and downsizing the ubiquitous role of grounding for meaning. In fact, there is growing support in favor of a form of *representational pluralism* (Dove 2009, Binder 2016), according to which all concepts consist of experiential and linguistic representations, while differing for the relative salience of these components. Dove (2023) argues for a *linguistic embodiment hypothesis*, according to which concepts in part rely on simulations of linguistic experiences. Therefore, linguistic co-occurrences, like the ones used by foundation models, do shape our meanings, alongside other types of multimodal experiences (Lenci 2008, Lupyan and Lewis 2019). Statistical distributions extracted from linguistic contexts are claimed to play a major role in the acquisition of abstract words (Vigliocco, Meteyard, Andrews and Kousta 2009, Lupyan 2019), or to explain how visual properties can be learned by congenitally blind people (Lewis, Zettersten and Lupyan 2019). Thus, we could argue that, although they lack referential knowledge, text-only foundation models have *some* «knowledge» of meaning (Piantadosi and Hill 2022), in the same sense in which I can know *something* of the meaning of *aardvark* without being able to point at its referents. Whether even this more limited statement is really correct will be discussed in Section 3.2.

A second critique to foundational models is what I will refer to as the *systematicity argument*. While the grounding argument mainly targets the kind of data these systems acquire their knowledge from, the systematicity one essentially concerns the type of generalizations they learn. A key property of human linguistic competence is *productivity*, that is the possibility of interpreting and generating a potentially infinite number of expressions. This entails the capacity of inducing generalizations that allow subjects to project their linguistic abilities beyond the data they are exposed to. As illustrated in Section 1, *prima facie* the last generation of neural NLUs seems to possess human-analogous productivity, which in turn would suggest they have been able to acquire the general rules and principles governing natural language. This point is instead denied by Berent and Marcus (2019), who claim that neural networks, and therefore all the complex architectures based on them, lack a key aspect of human cognition: The ability of carrying out *across-the board-generalizations*, that is «generalizations to any member of a category, irrespective of its similarity to training items» (p. e80). According to them, human learning mechanisms include the capacity to form abstract categories that treat all of their members alike and to operate over such classes algebraically using variables. On the other hand, neural models *do* generalize to new items, but only as long as they are similar to training data. Therefore, their productivity would be limited to *analogical generalizations*, while across-the-board ones are not similarity-driven, since algebraic rules can be applied to any new item that can fill their variables. This would prevent neural models to learn *systematic generalizations*, which Berendt and Marcus regard as a central aspect of natural language.

The systematicity argument is substantially a variation of the one advanced by Fodor and Pylyshyn (1988) against connectionist models, as the generation of neural networks in the 1980s was called. According to them human cognition and linguistic competence are characterized by the property of *systematicity*: The ability to produce/understand some sentences is intrinsically connected to the ability to produce/understand certain others. This in turn presupposes the other central property of *compositionality*: A lexical item must make approximately the same contribution to each expression in which it occurs. For instance, (4a) and (4b) are systematically related, because whoever understands the former must also understand the latter:

(4)     a. *The black cat chases the brown dog.*
         b. *The black dog chases the brown cat.*

This systematic relation depends on the fact that the sentences are instances of the same general structures licensed by English syntax, and that the words have the same meaning in the two sentences. If we fail to identify their systematic relations, it means that we have not mastered the across-the-board-generalizations behind them.

Like Fodor and Pylyshyn (1988), Berent and Marcus (2019) argue that systematicity and across-the-board-generalizations can be explained only by systems, like symbolic ones, that combine internally structured representations with variables. Since neural networks represent information with vectors that lack any internal structure, they would lack systematicity, and therefore they could not match a key aspect of human cognition. For instance, a neural network that produces two distinct embeddings for (4a) and (4b), might capture some aspects of their content, but nevertheless would not account for their systematic relations and the fact that the two sentences are instances of the same general structure that could generate potentially unlimited, systematically related sentences. Berent and Marcus (2019) essentially argue that even the last generation of neural networks grounding foundation models still suffer of the same problem, exactly because they lack those formal structures with variables that would be essential to capture systematicity and compositionality. In fact, this argument is supported by experiments that reveal that the most recent neural models trained on simplified or artificial language data show generalization abilities but fail to generalize in a systematic way (Lake and Baroni 2018, Goodwin, Sinha and Donnell 2020, Hupkes, Dankers, Mul and Bruni 2020). Pedinotti *et al.* (2021) and Kauf *et al.* (2022) show that foundations models acquire knowledge about events and their plausible participants (e.g., that a cop arresting a thief is a more plausible event than a thief arresting a cop), but this is often very dependent upon specific lexical patterns and lack the same generality as human one. The limits of the generalization abilities in large language models are also explored by Collins, Wong, Feng, Wei and Tenenbaum (2022) and Press *et al.* (2022).

The systematicity argument could be questioned for assuming compositionality, systematicity, and across-the-board generalizations as essential properties of natural language. Several linguistic phenomena suggest that natural language is at most *quasi-compositional* (Rabovsky and McClelland 2019). The hypothesis about the contextually invariant nature of lexical meaning contrasts with the sense modulation that lexical items constantly undergo when they are composed. In fact, the productivity that a semantic theory is called to explain concerns not only the ability of generating and understanding a potentially unlimited number of complex expressions, but also the ability of generating and understanding a potentially unlimited number of new word senses in context (Pustejovsky 1995). It is this context-sensitiveness that is captured by the contextual embeddings learned by foundations models (cf. Section 2). Moreover, the systematicity argument is weakened by the pervasiveness of nonsystematic and semiregular processes in language, which apply to categories of entities governed by overlapping and complex constraints (Johnson 2004). Rather than being the realm of across-the-board generalizations, natural language is characterized by the *partial productivity* (Goldberg 2019) of analogical generalizations based on the similarity to previously witnessed exemplars. Therefore, the fact that neural models strive to generalize compositionally and systematically does not entail they are unable to capture important aspects of natural language, since it also departs from these same properties (Baroni 2019). The quasi-compositional and partially productive nature of linguistic generalizations represent important challenges exactly for computational models based on categorical representations like the symbolic ones advocated by the proponents of the systematicity argument. The continuous representations of neural models are conversely more promising to address the aspects of natural language competence that are governed by analogical processes, gradience and similarity.

3.2. WHAT IS STILL MISSING?

The arguments we have reviewed in the previous section correctly emphasize crucial aspects of the human knowledge of language that state-of-the-art NLU systems still lack, despite their much-improved abilities. However, there are two equally important facts that must be highlighted: *The huge extent of what they are able to learn from textual data through the simple language modeling training objective, and the central role that prediction-based distributional learning has in human cognition too*. In fact, Goldstein *et al.* (2022) claim that autoregressive language models like GPT share with the brain important computational principles, since the brain is also constantly involved in next-word prediction as it processes natural language and represents words with contextual embeddings that incorporate several syntactic, semantic, and pragmatic properties of linguistic contexts. The effectiveness of distributional learning as a knowledge induction mechanism depends on the fact that language has evolved to communicate our experiences about the world and encodes relevant aspects of such experiences in its structures, as argued by the *Symbol Interdependence Hypothesis* proposed by Louwerse (2011). These experiences include embodied dimensions (e.g., visual, spatial, affective, etc.), but also pragmatic and social ones (Andreas 2022), which can thus be recovered from co-occurrence data. Using a prompting method, Hu *et al.* (2022) systematically investigate the pragmatic abilities of foundation models (e.g., recognizing indirect speech acts, understanding metaphors and irony, etc.) and find that they solve *some of them* with an accuracy close to human one. Correctly, they argue that their experiments do not show that neural language models have «pragmatic understanding», but that «explicit mentalizing is not *necessary* to mimic pragmatic behaviors - instead, experience with linguistic forms may be sufficient for deriving many human-like behavioral patterns». This is consistent with cognitive evidence indicating that language understanding does not always consists in the construction of full-fledged, highly structured semantic representations or complex reasoning processes. According to *good-enough models* of sentences processing, subjects use simple surface heuristics to build representations that are «good enough» to tackle a certain language understanding task (Karimi and Ferreira 2016). These cues act as «shortcuts» allowing subjects to process language quickly and achieve their goals efficiently. Such heuristics may well include distributional cues strongly associated with specific communicative intents or semantic dimensions, like the ones that neural language models are so good to learn from the linguistic input and exploit to solve communication tasks.

Therefore, on the one hand the «magic» of foundational models is simply the «magic» of distributional learning. The real scientific revelation brought by these models is that the range of semantic aspects that language encodes and can be recovered from distributional statistics is far greater than we could have ever imagined before (at least if we have enough amount of data). On the other hand, if a system that solves a semantic task only by using surface co-occurrences should be considered a «stochastic parrot», then we should admit that humans too often behave like «stochastic parrots», because they also *can* understand language using shallow surface cues only. From this perspective, language «understanding» in machines is similar to *some forms* of language understanding in humans, the extension of which is an interesting empirical problem that neural language models can help investigate. However, if humans too *can and do* resort to surface, distributional shortcuts to carry out language understanding, they do not *necessarily* do that, because there is something more in the human knowledge of language and its use in communication task than co-occurrence-based learning and processing. So, the real issue is: *What is missing in the way foundations models acquire, represent, and use their «core knowledge»?*

First, not all aspects of knowledge are expressed in language and therefore can be recovered from it. This is an effect of the so-called *reporting bias* (Gordon and Van Durme 2013), according to which information that speakers assume to share with the listeners is likely to be omitted from their linguistic productions. In turn, the reporting bias is a consequence of the maxim of quantity postulated by Grice (1975), which states that communication should be as informative as necessary, but no more, leaving unstated information that can be expected to be known. For example, the corpus analysis in Paik, Aroca-Ouellette, Roncone and Kann (2021) shows that color information about concepts associated with a single color (like strawberry) is worst represented. Zhang, Van Durme, Li and Stengel-Eskin (2022) find that neural language models have limited knowledge of the typical visual attributes of objects (e.g., their shape), and that bigger models show little or no improvement. However, this problem might just be a contingent limit due to the lack of grounding of text-only foundation models, which could be solved by giving them access to extralinguistic information, as multimodal ones have.[11] In fact, what is really missing in foundation models is not specific information unavailable in the type of data they are trained on: *They lack the ability to represent and organize what they acquire from texts through distributional learning into proper knowledge structures and to use such structures to solve language understanding tasks*.

In section 2, I said that one could argue that foundation models at least have inferential competences, in the sense of Marconi (1997). However, this does not seem the case either. Mahowald *et al.* (2023) distinguish the following major kinds of inferential knowledge: i.) *formal reasoning*, such as logical reasoning and novel problem solving, ii.) *world knowledge*, that is knowledge of objects and events and their properties, participants and relations; iii.) *situation modeling,* as the ability of building a representation of the stories we extract from language input and track their dynamic evolution over time; iv.) *social reasoning*, as the ability of using language by taking into account the states of mind of our interlocutors and our shared knowledge. Mahowald *et al.* (2023) review evidence revealing that in all these areas even the largest models like GPT-3 show performances still very far from human ones (Helwe, Clavel and Suchanek 2021, Kauf *et al.* 2022, Schuster and Linzen 2022, among several others). I think the major cause of these limits resides in the organization itself of the semantic spaces that foundation models extract from distributional data.

Inspired by *Conceptual Role Semantics* (Harman 1982), Piantadosi and Hill (2022) argue that neural language models do learn meaning, if this is conceived as a role within a *structured conceptual domain* or *«theory»* (cf. Section 2). The problem is that the semantic spaces learned by foundations models, like all other DSMs, are mainly arranged by *association* and *similarity*. At most they show some organization in terms of semantic categories but *lack more complex and articulated conceptual structure*. Contextual embeddings improve the representation of word senses, but on the other hand show the very same limit of former types of static distributional representations. Embeddings do encode several semantic properties, but the organization of the semantic space of foundations models is far from being a structured domain. Since the first explorations by Mikolov, Yih and Zweig (2013), relational structures have been argued to be recoverable from the geometry of embedding spaces (e.g., from vector differences). Unfortunately, further research has shown that this seems to be limited to few cases, which do not include sophisticated semantic structures (Lenci and Sahlgren 2023).

The semantic behavior of foundation models is determined by their internal representations, that is the information encoded in their embeddings. Consider now this example:

(5) Human question: *A tree in my garden is 60 years old. Last year its height was 180 cm. Then, it has gained 20 cm. How tall is the tree now?*
GPT-3: *The tree is now 200 cm tall.*

The system's answer is mathematically correct and fully coherent with respect to the question. Take now this other case:

(6) Human question: *My friend John is 60 years old. Last year his height was 180 cm. Then, he has gained 20 cm. How tall is he now?*
GPT-3: *John is now 200 cm tall.*

---

[11] These are in fact reported to have better performances in the identification of colors and other visual properties (Paik *et al.* 2021, Zhang *et al.* 2022).

The system's answer is again mathematically correct, but it is pragmatically odd. The problem is that the question itself does not make sense, because we know that, differently from trees, the height of humans cannot grow when they are sixty years old. The correct – intelligent – answer would have been to reply that the question cannot really be answered or that the situation illustrated in the question is impossible. Distinguishing the two cases requires having a «theory» about trees, humans and how their heights change. Either GPT-3 has embeddings that do not encode such a «theory», or it does not use it to generate its answers. The system's behavior seems instead to be guided by more surface patterns that prompt it to do the calculation, independently of whether doing it really makes sense or not. Therefore, superficially the behavior of GPT-3 is perfectly correct, but this correctness hides its lack of any deeper language understanding.

The following dialogue with GPT-3 is also illuminating about the difference between having acquired a distributional embedding semantic space and a real knowledge structure:

(7) Human question: *Who was Cicero?*
GPT-3: *Cicero was a Roman statesman, lawyer, scholar, and philosopher who lived from 106 BC to 43 BC. He was one of the most important figures of the Roman Republic, and his writings were highly influential in the development of Roman law. He is also known for his orations, particularly the famous Catiline Orations, in which he denounced Catiline for his alleged conspiracy against the Roman government.*

Human question: *When tobacco was discovered?*
GPT-3: *Tobacco was first discovered by the native peoples of the Americas. It is believed that the use of tobacco dates back as far as 5,000-7,000 years ago, and it was used for both medicinal and ceremonial purposes. Tobacco was introduced to Europe by Christopher Columbus in 1492, and it quickly became popular throughout the continent.*

Human question: *What did Cicero smoke?*
GPT-3: *Cicero did not smoke, as smoking was not a common practice during his lifetime in the 1st century BC. He did, however, write extensively on the subject of tobacco, which he viewed as a dangerous and unhealthy habit. In his writings, he warned of the dangers of smoking and urged people to avoid it.*

This dialogue shows that GPT-3 «knows» a lot about Cicero and tobacco, but this «knowledge» cannot prevent it from stating that Cicero wrote about tobacco and its dangerous effects.[12] These are the cases that indeed make the model resemble a «stochastic parrot». The point is that such behavior does not depend on its lack of grounding, but on the lack of a properly structured knowledge space.

Reasoning like the one in (7) presuppose the availability of «theories» and the ability of using them to solve language understanding and generation tasks. The stories we extract from novels or movies also have a rich conceptual structure, with complex hierarchical relations between characters and events (Cohn, Jackendoff, Holcomb, Kuperberg 2014). Schuster and Linzen (2022) show that GPT-3 is not even able to distinguish cases like *John bought a dog* in which the indefinite noun phrase introduces a new discourse entity to which we can anaphorically refer later on (e.g., *it is a golden retriever*), from cases in which the presence of a negation (i.e., *John didn't buy a dog*) blocks the introduction of a new referent into the discourse model. The lack of «theories» also affects social reasoning, which requires «theories of mind», that is the ability to reason about the others' mental states such as beliefs and intentions (Enrici, Bara and Adenzato 2019). Mutual recursive reasoning is an essential part of human communication (Grice 1975): A speaker A generates an utterance keeping in mind how a listener B is likely to interpret it, and B interprets the utterance using hypotheses about the possible intents and knowledge states of A, recursively assuming that A has produced it under some assumptions about B's mind. It is this reasoning about each other «theories of mind» that allow speaker and listener to arrive at the «tacit coordination» (Levinson 2000) that is the key for successful communication. A lot of pragmatic aspects are strongly conventionalized and encoded in linguistic structures (Rubio-Fernández 2021). This allows speakers to use shallow heuristics as «shortcuts» to deep pragmatic reasoning. But this is not always the case, and deeper reasoning is often necessary to reach our communication goals. For instance, the experiments in Hu *et al.* (2022) show that in cases like irony or humor the behavior of foundations models, which only use distributional information, is almost at chance level. Sap, LeBras, Fried and Choi (2022) also report that the ability of neural language models in tasks designed to test their social intelligence is very far from the human one.

---

[12] GPT-3 text generation is stochastic, and therefore the system can generate different answers to the same question each time it is prompted. The system's answer can be appropriate (e.g., it says that Cicero could not smoke because Romans did not know tobacco), but this is just an exception among most answers that are like (7). Therefore, even the correct response looks like a mere accident, which again confirms that it is just the product of shallow statistic associations, rather than deep semantic reasoning.

In summary, I believe the major problem of state-of-the art NLU systems based on foundation models concerns the very type of «knowledge» they extract from linguistic data. They mainly identify highly sophisticated associative links between linguistic expressions, but they do not have a semantic space organized in terms of structured «theories» that might support a truly inferential competence. They do not have «theories of the world», let alone «theories of mind», which therefore prevents them from developing those advanced pragmatic skills that are essential for language use. In other terms, the «core knowledge» of foundation models is rich of factoids and associations (far more than any human being could ever master, given the huge amount of data they are extracted from), but it lacks the same structured organization as the human knowledge system. This cannot be simply solved by grounding models in the external world. In fact, multimodal foundation models adopt the same prediction-based mechanism as text-only ones. Though they learn associations between language and the world, besides associations between linguistic forms, still they do not learn «theories of the world». The lack of proper knowledge structures is also related to the limited generalization abilities of large language models. Quasi-regularity is pervasive in natural language, analogical and similarity-driven inferences are also key aspects of human cognition. But it is equally true that several aspects of human linguistic and non-linguistic intelligence depend on fully abstract inference schemas, akin to the across-the-board generalizations advocated by Berent and Marcus (2019), which foundation models strive to perform. Therefore, their «knowledge» is very different from our knowledge under several respects, even if we consider its inferential component only. Statistical associations and similarity can complement inferential reasoning but cannot replace it.

4. CONCLUSIONS

Floridi and Chiriatti (2020) claim that «the real point about AI is that we are increasingly decoupling the ability to solve a problem effectively – as regards the final goal – from any need to be intelligent to do so» (p. 683). I think this is a correct and faithful analysis of what is happening. Current NLU systems based on foundation models have reached an unprecedented ability of «mimicking» human linguistic behavior, without having the same intelligence that humans use when they use language. They solve many linguistic tasks only relying on statistical regularities extracted by neural networks optimized to predict strings. What these models can do with such a simple mechanism – though implemented by gigantic architectures trained on huge amounts of texts – is indeed extraordinary, and up to a certain point approximate cognitively plausible mechanisms of language learning and processing. However, these models are nothing else but very clever illusionists. They give us the illusion of «talking like us», an illusion that, we must admit, is nowadays extremely realistic, so that they have *de facto* made the classical *Turing Test* based on the «imitation game» obsolete or at least ineffective (Elkins and Chun 2020). We, the amazed audience, are sometimes able to spot the trick from tiny imperfections in the show. Bigger models will make the illusion even more perfect and will be integrated in thousands of applications (with positive and negative effects, as usual), but this will not change what they really are, just linguistic illusionists.

Human speakers often play the same tricks. Surface heuristics are powerful «shortcuts» to more complex inference patterns and allow speakers to exploit what is encoded and conventionalized in language to speed up its processing and play «communication games» more effectively. Statistical distributional learning has a central role in cognition and is probably more powerful that it has been assumed before, though current models need unrealistic quantities of data to learn linguistic structures. However, human natural language understanding requires much more than this. Mahowald *et al.* (2023) claim that large language models at most have *formal linguistic competence*, that is knowledge of linguistic rules and patterns, while they fall short of *functional competence*, as the ability of understanding and using language in the world. The reason is that the latter requires complex «theories of the world and mind» and mechanisms that allow their use to drive linguistic behavior. This is a type of information that is still qualitatively different from the one that foundation models seem to be able to acquire from linguistic data and to represent in their continuous embeddings.

What is then the road in front of us? What might help us reach the coveted Grail? Surely it is not only a matter of making the models bigger. First, we must increase our ability to spot the trick in their illusions. Since the mere observational adequacy of linguistic production is not enough, solid scientific advancement will only be achieved through fine-grained tests on carefully designed datasets that can give us reliable evidence of models' ability to account for specific aspects of natural language understanding (one recent example is *BIG-bench, Beyond the Imitation Game benchmark*; Srivastava *et al.* 2022). Moreover, we must explore other tracks to «climb the hill», especially those that can overcome the classical dichotomy between symbolic and neural models that is still alive and dominant today, in both sides of the barricade. It is instead likely that new breakthrough towards genuine human-like language understanding may come from integrating neural language models with structured knowledge that will combine the flexibility of data-driven continuous representations with the inferential power of symbolic systems.

ACKNOWLEDGEMENTS

This research was partly funded by PNRR - M4C2 - Investimento 1.3, Partenariato Esteso PE00000013 – «FAIR - Future Artificial Intelligence Research» - Spoke 1 «Human-centered AI», funded by the European Commission under the NextGeneration EU programme.